\documentclass[letterpaper]{article} 
\usepackage{aaai24} 
\usepackage[table]{xcolor} 

\usepackage{times}  
\usepackage{helvet}  
\usepackage{courier}  
\usepackage[hyphens]{url}  
\usepackage{graphicx} 
\usepackage{natbib}  
\usepackage{caption} 
\usepackage{algorithm}
\usepackage[T1]{fontenc}
\usepackage{epsfig}
\usepackage{amsmath}
\usepackage{amssymb}
\usepackage{booktabs}
\usepackage{bbding}
\usepackage{pifont}
\usepackage{wasysym}
\usepackage{utfsym}
\usepackage{fontawesome}
\usepackage{color}
\usepackage{algorithmicx}  
\usepackage{algpseudocode}
\usepackage{comment}
\usepackage{tablefootnote}
\usepackage{multirow}
\usepackage{subcaption}

\newcommand{\far}{Far3D}

\usepackage{newfloat}
\usepackage{listings}
\DeclareCaptionStyle{ruled}{labelfont=normalfont,labelsep=colon,strut=off} 
\floatstyle{ruled}
\newfloat{listing}{tb}{lst}{}
\floatname{listing}{Listing}
\title{Far3D: Expanding the Horizon for Surround-view 3D Object Detection}
\author{
    Xiaohui Jiang $^{*1\dagger}$ \hspace{0.25cm} Shuailin Li $^{*2}$ \hspace{0.25cm} Yingfei Liu$^2$ \hspace{0.25cm} Shihao Wang$^{1\dagger}$ \hspace{0.25cm} Fan Jia$^2$ \hspace{0.25cm} Tiancai Wang$^2$ \hspace{0.25cm} Lijin Han$^{1}$  \hspace{0.25cm}  Xiangyu Zhang$^2$ \\ 
}
\affiliations{
    $^1$Beijing Institute of Technology \hspace{0.9cm} $^2$MEGVII Technology \\

}

\usepackage{bibentry}

\begin{document}

\maketitle
\let\thefootnote\relax\footnotetext{* Equal contribution.} 
\let\thefootnote\relax\footnotetext{$\dagger$ Work done during the internship at MEGVII Technology. } 
\begin{abstract}
Recently 3D object detection from surround-view images has made notable advancements with its low deployment cost. However, most works have primarily focused on close perception range while leaving long-range detection less explored.
Expanding existing methods directly to cover long distances poses challenges such as heavy computation costs and unstable convergence. 
To address these limitations, this paper proposes a novel sparse query-based framework, dubbed \far{}. By utilizing high-quality 2D object priors, we generate 3D adaptive queries that complement the 3D global queries.
To efficiently capture discriminative features across different views and scales for long-range objects, we introduce a perspective-aware aggregation module. Additionally, we propose a range-modulated 3D denoising approach to address query error propagation and mitigate convergence issues in long-range tasks.
Significantly, \far{} demonstrates SoTA performance on the challenging Argoverse 2 dataset, covering a wide range of 150 meters, surpassing several LiDAR-based approaches. 
The code is available at https://github.com/megvii-research/Far3D.


    
\end{abstract}
 \section{Introduction}

3D object detection plays an important role in understanding 3D scenes for autonomous driving, aiming to provide accurate object localization and category around the ego vehicle. Surround-view methods~\cite{huang2022bevdet4d, li2023bevdepth, liu2022petrv2,  li2022bevformer, yang2023bevformer, park2022time, wang2023exploring}, 
with their advantages of low cost and wide applicability, have achieved remarkable progress. 
However, most of them focus on close-range perception (e.g., $\sim$50 meters on nuScenes~\cite{caesar2020nuscenes}), leaving the long-range detection field less explored. Detecting distant objects is essential for real-world driving to maintain a safe distance, especially at high speeds or complex road conditions.

\begin{figure}[t!]
    \centering  
    \includegraphics[width=0.49\textwidth]{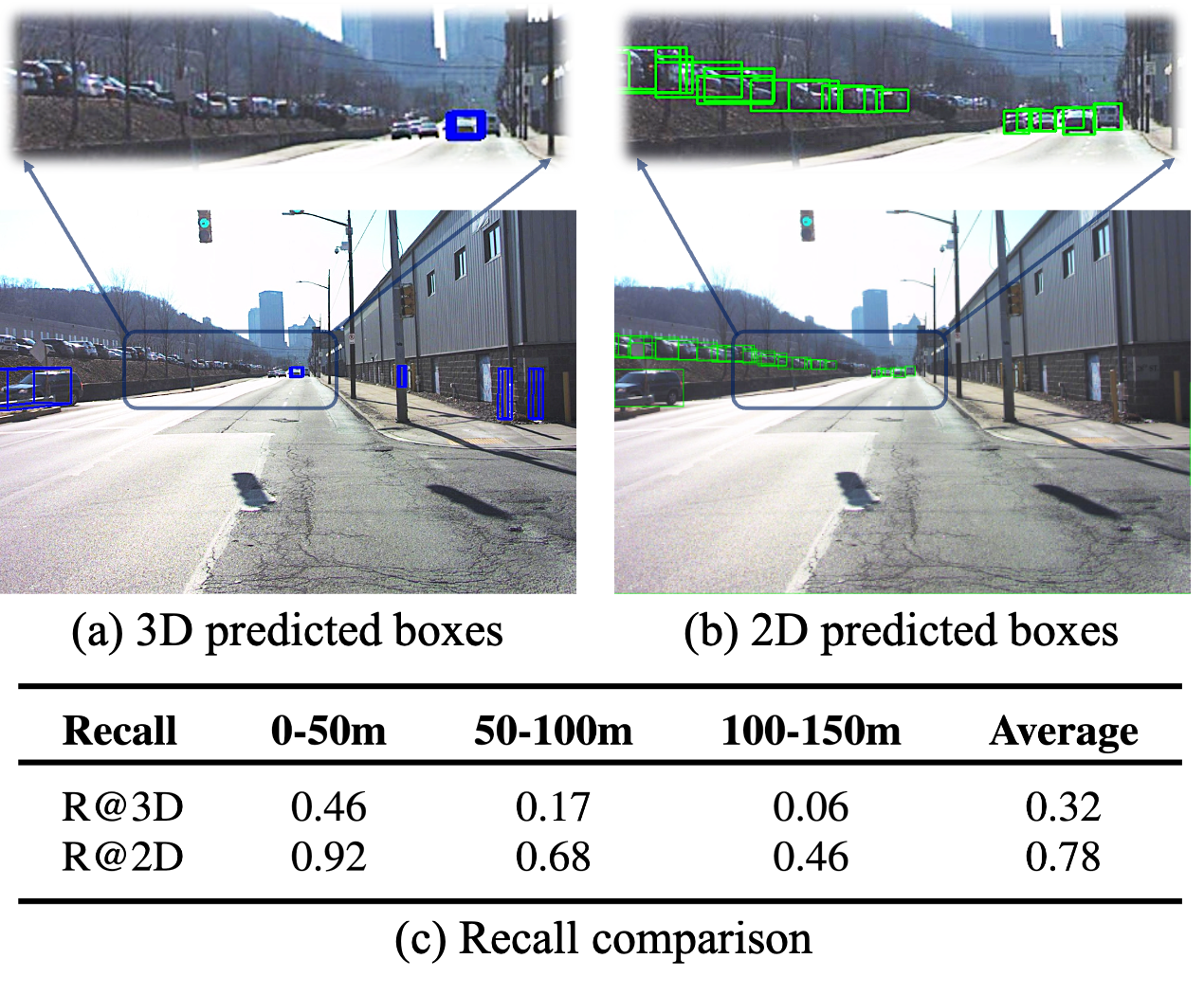}
    \caption{Peformance comparisons on Argoverse 2 between 3D detection and 2D detection. (a) and (b) demonstrate predicted boxes of StreamPETR and YOLOX, respectively. (c) imply that 2D recall is notably better than 3D recall and can act as a bridge to achieve high-quality 3D detection. Note that 2D recall does not represent 3D upper bound due to different recall criteria.}
    \label{fig:intro}
\vspace{-0.5cm}
\end{figure}

Existing surround-view methods can be broadly categorized into two groups based on the intermediate representation, dense Bird's-Eye-View~(BEV) based methods and sparse query-based methods. BEV based methods~\cite{huang2021bevdet, huang2022bevdet4d, li2023bevdepth,li2022bevformer, yang2023bevformer} usually convert perspective features to BEV features by employing a view transformer~\cite{philion2020lift}, then utilizing a 3D detector head to produce the 3D bounding boxes. However, dense BEV features come at the cost of high computation even for the close-range perception, making it more difficult to scale up to long-range perception.
Instead, following DETR~\cite{carion2020end} style, sparse query-based methods~\cite{wang2022detr3d, liu2022petr, liu2022petrv2, wang2023exploring} adopt learnable global queries to represent 3D objects, and interact with surround-view image features to update queries.
Although sparse design can avoid the squared growth of query numbers, its global fixed queries cannot adapt to dynamic scenarios and usually miss targets in long-range detection.
We adopt the sparse query design to maintain detection efficiency and introduce 3D adaptive queries to address the inflexibility weaknesses.

\begin{figure}[t]
\centering
\includegraphics[scale=0.17]{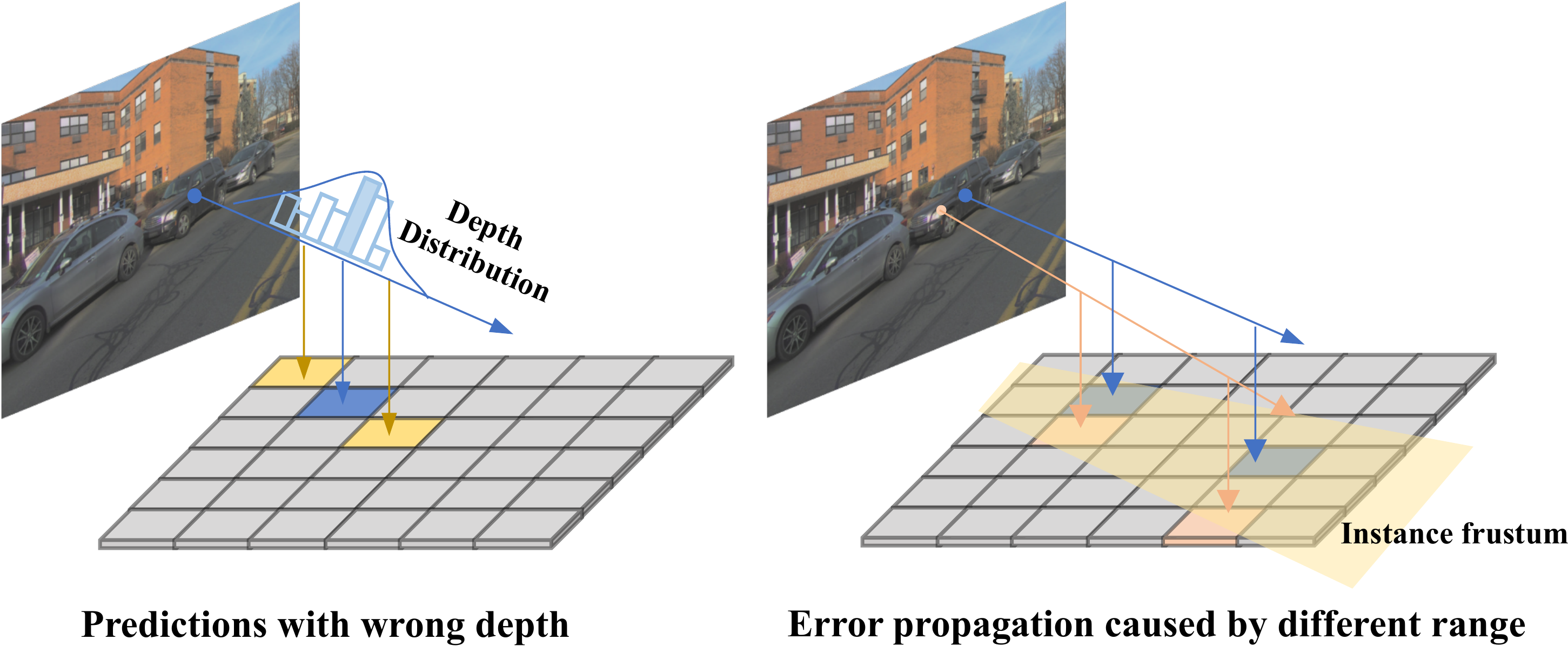}
\caption{Different cases of transformimg 2D points into 3D space. The blue dots indicate the centers of 3D objects in images. (a) shows the redundant prediction with the wrong depth, which is in yellow. (b) illustrates the error propagation problem dominated by different ranges.}
\label{fig:dn}
\vspace{-0.5cm}
\end{figure}

To employ the sparse query-based paradigm for long-range detection, the primary challenge lies in poor recall performance. Due to the query sparsity in 3D space, assignments between predictions and ground-truth objects are affected, generating only a small amount of matched positive queries. 
As illustrated in Fig.~\ref{fig:intro}, 3D detector recalls are pretty low, yet recalls from the existing 2D detector are much higher, showing a significant performance gap between them. Motivated by this, leveraging high-quality 2D object priors to improve 3D proposals is a promising approach, for enabling accurate localization and comprehensive coverage. Although previous methods like SimMOD~\cite{zhang2023simple} and MV2D~\cite{wang2023object} have explored using 2D predictions to initialize 3D object proposals, they primarily focus on close-range tasks and discard learnable object queries. 
Moreover, as depicted in Fig.~\ref{fig:dn}, directly introducing 3D queries derived from 2D proposals for long-range tasks encounters two issues: 1) inferior redundant predictions due to uncertain depth distribution along the object rays, and 2) larger deviations in 3D space as the range increases due to frustum transformation. These noisy queries can impact the training stability, requiring effective denoising ways to optimize. Furthermore, within the training process, the model exhibits a tendency to overfit on densely populated close objects while disregarding sparsely distributed distant objects.


To address the aforementioned challenges, we design a novel 3D detection paradigm to expand the perception horizon. Despite the 3D global query that was learned from the dataset, our approach also incorporates auxiliary 2D proposals into 3D adaptive query generation. 
Specifically, we first produce reliable pairs of 2D object proposals and corresponding depths then project them to 3D proposals via spatial transformation. We compose 3D adaptive queries with the projected positional embedding and semantic context, which would be refined in the subsequent decoder.
In the decoder layers, perspective-aware aggregation is employed across different image scales and views. It learns sampling offsets for each query and dynamically enables interactions with favorable features. For instance, distant object queries are beneficial to attend large-resolution features, while the opposite is better for close objects in order to capture high-level context.
Lastly, we design a range-modulated 3D denoising technique to mitigate query error propagation and slow convergence. Considering the different regression difficulties for various ranges, noisy queries are constructed based on ground-truth (GT) as well as referring to their distances and scales. Our method feeds multi-group noisy proposals around GT into the decoder and trains the model to a) recover 3D GT for positive ones and b) reject negative ones, respectively. The inclusion of query denoising also alleviates the problem of range-level unbalanced distribution.

Our proposed method achieves remarkable performance advancements over state-of-the-art (SoTA) approaches in the challenging long-range Argoverse 2 dataset, as well as surpassing the prior arts of LiDAR-based methods. 
To evaluate the generalization capability, we further validate its results on nuScenes dataset and demonstrate SoTA metrics. 

In summary, our contributions are:
\begin{itemize}
\item We propose a novel sparse query-based framework to expand the perception range in 3D detection, by incorporating high-quality 2D object priors into 3D adaptive queries.
\item We develop perspective-aware aggregation that captures informative features from diverse scales and views, as well as a range-modulated 3D denoising technique to address query error propagation and convergence problems.
\item On the challenging long-range Argoverse 2 datasets, our method surpasses surround-view  methods and 
outperforms several LiDAR-based methods. The generalization of our method is validated on the nuScenes dataset. 

\end{itemize} 


\section{Related Work}

\subsection{Surround-view 3D Object Detection}
Recently 3D object detection from surround-view images has attracted much attention and achieved great progress, due to its advantages of low deployment cost and rich semantic information. 
Based on feature representation, existing methods~\cite{wang2021fcos3d, wang2022detr3d, liu2022petr, huang2022bevdet4d, li2023bevdepth, li2022bevstereo, jiang2023polarformer, liu2022petrv2,  li2022bevformer, yang2023bevformer, park2022time, wang2023exploring, zong2023temporal, liu2023towards} can be largely classified into BEV-based methods and sparse-query based methods.  
\begin{figure*}[t]
\centering
\includegraphics[scale=0.17]{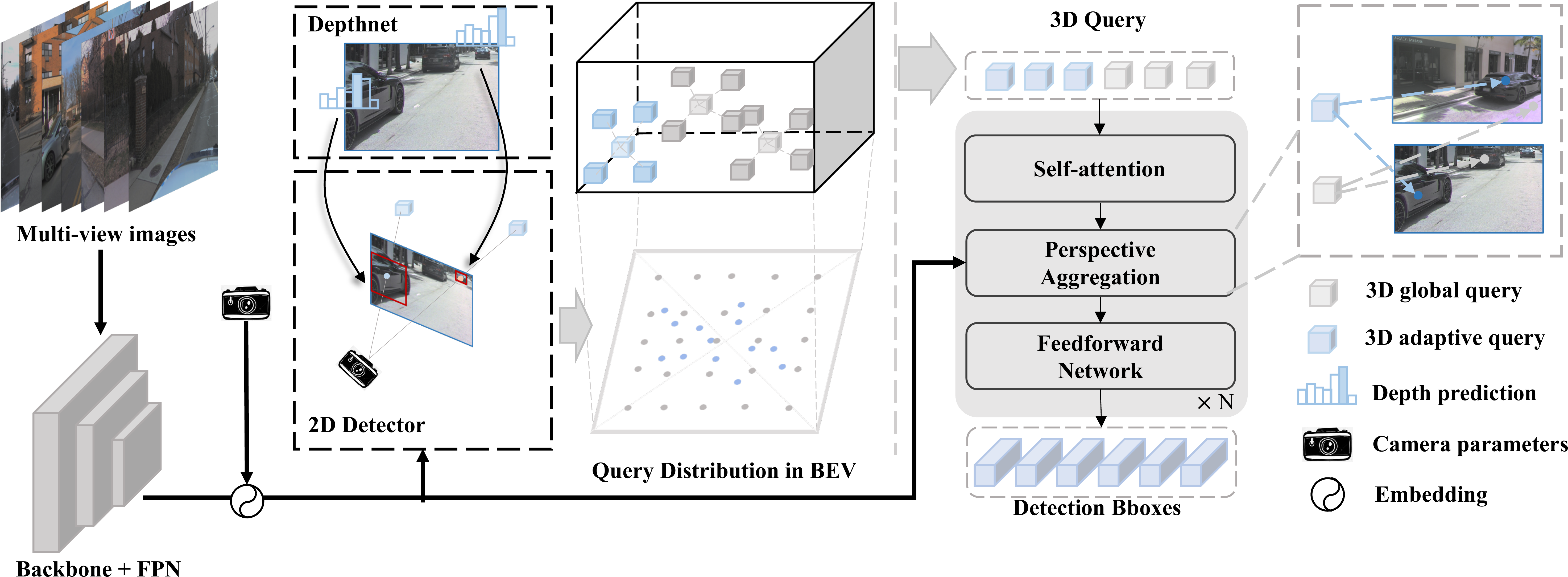}
\caption{The overview of our proposed Far3D. 
Feeding surround-view images into the backbone and FPN neck, we obtain 2D image features and encode them with camera parameters for perspective-aware transformation. 
Utilizing a 2D detector and DepthNet, we generate reliable 2D box proposals and their corresponding depths, which are then concatenated and projected into 3D space. 
The generated 3D adaptive queries, combined with the initial 3D global queries, are iteratively refined by the decoder layers to predict 3D bounding boxes. Furthermore, temporal modeling is equipped through long-term query propagation.
}
\label{fig:method_overall}
\vspace{-0.3cm}
\end{figure*}

Extracting image features from surround views, BEV-based methods~\cite{huang2021bevdet, huang2022bevdet4d, li2023bevdepth, li2022bevformer} transform features into BEV space by leveraging estimated depths or attention layers, then a 3D detector head is employed to predict localization and other properties of 3D objects. 
For instance, 
BEVFormer~\cite{li2022bevformer} leverages both spatial and temporal
features by interacting with spatial and temporal space through predefined grid-shaped BEV queries.
BEVDepth~\cite{li2023bevdepth} propose a 3D detector with a trustworthy depth estimation, by introducing a camera-aware depth estimation module.
On the other hand, sparse query-based paradigms~\cite{wang2022detr3d, liu2022petr} learn global object queries from the representative data, then feed them into the decoder to predict 3D bounding boxes during inference. This line of work has the advantage of lightweight computing. 

Furthermore, temporal modeling for surround-view 3D detection can improve detection performance and decrease velocity errors significantly, and many works~\cite{huang2022bevdet4d, liu2022petrv2, park2022time, wang2023exploring, lin2022sparse4d, lin2023sparse4dv2} aim to extend a single-frame framework to multi-frame design. 
BEVDet4D~\cite{huang2022bevdet4d} lifts the BEVDet paradigm from the spatial-only 3D space to the spatial-temporal 4D space, via fusing features with the previous frame.
PETRv2~\cite{liu2022petrv2} extends the 3D position embedding in PETR for temporal modeling through the temporal alignment of different frames. 
However, they use only limited history.
To leverage both short-term and long-term history, SOLOFusion~\cite{park2022time} balances the impacts of spatial resolution and temporal difference on localization potential, then use it to design a powerful temporal 3D detector.
StreamPETR~\cite{wang2023exploring} develops an object-centric temporal mechanism in an online manner, where long-term historical information is propagated through object queries.

\subsection{2D Auxiliary Tasks for 3D Detection}
3D detection from surround-view images can be improved through 2D auxiliary tasks, and some works~\cite{xie2022m, zhang2023simple, wang2022focal, yang2023bevformer, wang2023object} aim to exploit its potential. There are several approaches including 2D pertaining, auxiliary supervision, and proposal generation. 
SimMOD~\cite{zhang2023simple} exploits sample-wise object proposals and designs a two-stage training manner, where perspective object proposals are generated and followed by iterative refinement in DETR3D-style.
Focal-PETR~\cite{wang2022focal} performs 2D object supervision to adaptively focus the attention of 3D queries on discriminative foreground regions.
BEVFormerV2~\cite{yang2023bevformer} presents a two-stage BEV detector where perspective proposals are fed into the BEV head for final predictions.
MV2D~\cite{wang2023object} designs a 3D detector head that is initialized by RoI regions of 2D predicted proposals.

Compared to the above methods, our framework differs in the following aspects. Firstly, we aim to resolve the challenges of long-range detection with surrounding views, which are less explored in previous methods.
Besides learning 3D global queries, we explicitly leverage 2D predicted boxes and depths to build 3D adaptive queries, utilizing positional prior and semantic context simultaneously. Furthermore, the designs of perspective-aware aggregation and 3D denoising are integrated to address task issues.
\section{Method}

\subsection{Overview}
Fig.~\ref{fig:method_overall} shows the overall pipeline of our sparse query-based framework. 
Feeding surround-view images $\mathbf{I}=\{\mathbf{I^1, ..., I^n}\}$, we extract  multi-level images features $\mathbf{F}=\{\mathbf{F^1, ..., F^n}\}$ by using the backbone network~(e.g. ResNet, ViT) and a FPN~\cite{lin2017feature} neck. 
To generate 3D adaptive queries, we first obtain 2D proposals and depths using a 2D detector head and depth network, then filter reliable ones and transform them into 3D space to generate 3D object queries. In this way, informative object priors from 2D detections are encoded into the 3D adaptive queries.

In the 3D detector head, we concatenate 3D adaptive queries and 3D global queries, then input them to transformer decoder layers including self-attention among queries and perspective-aware aggregation between queries and features. We propose perspective-aware aggregation to efficiently capture rich features in multiple views and scales by considering the projection of 3D objects. Besides, range-modulated 3D denoising is introduced to alleviate query error propagation and stabilize the convergence, when training with long-range and imbalanced distributed objects. Sec~\ref{sec:denoise} depicts the denoising technique in detail.

\subsection{Adaptive Query Generation}
Directly extend existing 3D detectors from short range (e.g. \textasciitilde50m) to long range (e.g. \textasciitilde150m) suffers from several problems: heavy computation costs, inefficient convergence and declining localization ability. For instance, the query number is supposed to grow at least squarely to cover possible objects in a larger range, yet such a computing disaster is unacceptable in realistic scenarios. Besides that, small and sparse distant objects would hinder the convergence and even hurt the localization of close objects. 
Motivated by the high performance of 2D proposals, we propose to generate adaptive queries as objects prior to assist 3D localization. This paradigm compensates for the weakness of global fixed query design and allows the detector to generate adaptive queries near the ground-truth (GT) boxes for different images. In this way, the model is equipped with better generalization and practicality.

Specifically, given image features after FPN neck, we feed them into the anchor-free detector head from YOLOX~\cite{ge2021yolox} and a light-weighted depth estimation net, outputting 2D box coordinates, scores and depth map.
2D detector head follows the original design, while the depth estimation is regarded as a classification task by discretizing the depth into bins~\cite{reading2021categorical, zhang2022monodetr}.
We then make pairs of 2D boxes and corresponding depths. To avoid the interference of low-quality proposals, we set a score threshold $\tau$~(e.g. 0.1) to leave only reliable ones. For each view $i$, box centers $(\mathbf{c}_w, \mathbf{c}_h)$ from 2D predictions and depth $\mathbf{d}_{wh}$ from depth map are combined and projected to 3D proposal centers $\mathbf{c}_{3d}$.
\begin{equation}
    \mathbf{c_{3d}} = \mathit{K_{i}^{-1}I_{i}^{-1}}[\mathbf{c}_w*\mathbf{d}_{wh}, \mathbf{c}_h*\mathbf{d}_{wh}, \mathbf{d}_{wh}, \mathbf{1}]^{T}
\end{equation}
where $\mathit{K_{i}, I_{i}}$ denote camera extrinsic and intrinsic matrices.

After obtaining projected 3D proposals, we encode them into 3D adaptive queries as follows,
\begin{equation}
    \mathbf{Q}_{pos} = \mathit{PosEmbed}(\mathbf{c}_{3d})
\end{equation}
\begin{equation}
    \mathbf{Q}_{sem} = \mathit{SemEmbed}(\mathbf{z}_{2d}, \mathbf{s}_{2d})
\end{equation}
\begin{equation}
    \mathbf{Q} = \mathbf{Q}_{pos} + \mathbf{Q}_{sem}
\end{equation}
where $\mathbf{Q}_{pos}, \mathbf{Q}_{sem}$ denote positional embedding and semantic embedding, respectively. $\mathbf{z}_{2d}$ sampled from $\mathbf{F}$ corresponds to the semantic context of position $(\mathbf{c}_w, \mathbf{c}_h)$, and $\mathbf{s}_{2d}$ is the confidence score of 2D boxes. $\mathit{PosEmbed(\cdot)}$ consists of a sinusoidal transformation~\cite{vaswani2017attention} and a MLP, while $\mathit{SemEmbed(\cdot)}$ is another MLP. 

Lastly, the proposed 3D adaptive queries are concatenated with initialized global queries, and fed to subsequent transformer layers in the decoder. 

\subsection{Perspective-aware Aggregation}
Existing sparse query-based approaches usually adopt one single-level feature map for computation effectiveness (e.g. StreamPETR). However, the single feature level is not optimal for all object queries of different ranges. For example, small distant objects require large-resolution features for precise localization, while high-level features are better suited for large close objects.
To overcome the limitation, we propose perspective-aware aggregation, enabling efficient feature interactions on different scales and views. 

Inspired by the deformable attention mechanism~\cite{zhu2020deformable}, we apply a 3D spatial deformable attention consisting of 3D offsets sampling followed by view transformation. Formally, we first equip image features $\mathbf{F}$ with the camera information including intrinsic $\mathbf{I}$ and extrinsic parameters $\mathbf{K}$.
A squeeze-and-excitation block~\cite{hu2018squeeze} is used to explicitly enrich the features. 
Given enhanced feature $\mathbf{F'}$, we employ 3D deformable attention instead of global attention in PETR series~\cite{liu2022petr, liu2022petrv2, wang2023exploring}. For each query reference point in 3D space, the model learns $M$ sampling offsets around and projects these references into different 2D scales and views.
\begin{equation} \label{eq:p_2d}
    \mathbf{P}_q^{2d} = \mathbf{I\cdot K \cdot (P}_q^{3d} + \Delta \mathbf{P}_q^{3d})
\end{equation}
where $\mathbf{P}_q^{3d}, \Delta\mathbf{P}_q^{3d}$ are 3D reference point and learned offsets for query $q$, respectively. $\mathbf{P}_q^{2d}$ stands for the projected 2d reference point of different scales and views. For simplicity, we omit the subscripts of scales and views.

\begin{table*}[th!]
\centering
\caption{Comparisons on the Argoverse 2 \texttt{val} set. We evaluate 26 object categories with a range of 150 meters. \far{} outperform previous surround-view methods with a large margin, and surpass several SoTA LiDAR-based methods. 
Surround-view methods except for PETR are with temporal modeling.
${}^\ddagger$ are reproduced by ourselves.
}
\vspace{-0.2cm}
\label{tab:av2_val_set}
\tiny
\resizebox{0.9\textwidth}{!}{
\setlength{\tabcolsep}{3.5pt}
\begin{tabular}{l|c|c|c|cc|cc@{\hspace{1.0\tabcolsep}}c@{\hspace{1.0\tabcolsep}}}  

\toprule
\textbf{Methods} & \textbf{Backbone} & \textbf{Modality} & \textbf{Image/Voxel Size}  & \textbf{mAP}$\uparrow$  &\textbf{CDS}$\uparrow$  & \textbf{mATE}$\downarrow$ & \textbf{mASE}$\downarrow$   &\textbf{mAOE}$\downarrow$    \\ 
\midrule
BEVStereo${}^\ddagger$ & VoV-99 & Camera & 960 $\times$ 640 
& 0.146 & 0.104 &0.847&0.397 & 0.901\\ 
SOLOFusion${}^\ddagger$ & VoV-99 & Camera & 960 $\times$ 640 
& 0.149 & 0.106 &0.934&0.425 & 0.779\\ 
PETR & VoV-99 & Camera & 960 $\times$ 640 
& 0.176 & 0.122 &0.911&0.339 & 0.819\\ 
Sparse4Dv2 & VoV-99 & Camera & 960 $\times$ 640 & 
0.189 & 0.134 & 0.832 & 0.343 & 0.723\\
StreamPETR & VoV-99 & Camera & 960 $\times$ 640           & 0.203 & 0.146 & 0.843 & 0.321 & 0.650 \\ 
\rowcolor[gray]{.9} 
\far{}~(Ours) & VoV-99 & Camera & 960 $\times$ 640                                 & \textbf{0.244} & \textbf{0.181} & \textbf{0.796} & \textbf{0.304} & \textbf{0.538} \\ 

\midrule 
\midrule 
CenterPoint & - & Lidar & (0.2, 0.2, 0.2) & 0.274 & 0.210 & 0.548 & 0.362 & 0.781 \\
FSD & - & Lidar & (0.2, 0.2, 0.2) & 0.291 & 0.233 & \textbf{0.468} & \textbf{0.299} & 0.740 \\
VoxelNeXt & - & Lidar & (0.1, 0.1, 0.2) & 0.307 & 0.225 & 0.431 & 0.291 & 1.157  \\ 
\rowcolor[gray]{.9} 
\far{}~(Ours) & ViT-L & Camera & 1536 $\times$ 1536                                  & \textbf{0.316} & \textbf{0.239} & 0.732 & 0.303 & \textbf{0.459} \\

\bottomrule
\end{tabular}}
\end{table*}  
\begin{table*}[th!]
\centering
\caption{Comparison on the nuScenes \texttt{val} and \texttt{test} splits. \far{} achieves the highest performance compared to prior-arts, validating its generalization ability. ${}^\ast$Benefited from the perspective-view pre-training. We employ the resolution 512 $\times$ 1408 for \texttt{val} and 1536 $\times$ 1536 for \texttt{test} split.
}
\vspace{-0.2cm}
\label{tab:nus_val_set}
\tiny
\resizebox{0.9\textwidth}{!}{
\setlength{\tabcolsep}{3.5pt}
\begin{tabular}{l|c|c|cc|ccccc}

\toprule
\textbf{Methods} & \textbf{Backbone} & \textbf{Split}  & \textbf{mAP}$\uparrow$  &\textbf{NDS}$\uparrow$  & \textbf{mATE}$\downarrow$ & \textbf{mASE}$\downarrow$   &\textbf{mAOE}$\downarrow$ &\textbf{mAVE}$\downarrow$   &\textbf{mAAE}$\downarrow$\\ 
\midrule
PETR & ResNet101 & \texttt{val}
& 0.366 & 0.441 & 0.717 & 0.261 & 0.412 & 0.834 & 0.190 \\ 
SOLOFusion & ResNet101 & \texttt{val}
& 0.483 & 0.582 & \textbf{0.503} & 0.264 & 0.381 & 0.246 & 0.207 \\ 
StreamPETR${}^\ast$ & ResNet101 & \texttt{val}
& 0.504 & 0.592 & 0.569 & 0.262 & \textbf{0.315} & 0.257 & 0.199 \\ 
Sparse4Dv2${}^\ast$  & ResNet101 & \texttt{val}
& 0.505 & 0.594 & 0.548 & 0.268 & 0.348 & 0.239 & \textbf{0.184} \\ 
\rowcolor[gray]{.9} 
\far{}~(Ours)${}^\ast$  & ResNet101 & \texttt{val}
& \textbf{0.510} & \textbf{0.594} & 0.551 & \textbf{0.258} & 0.372 & \textbf{0.238}& 0.195 \\ 
\midrule
\midrule
SOLOFusion & ConvNeXt-B & \texttt{test}
& 0.540 & 0.619 & 0.453 & 0.257 & 0.376 & 0.267 & 0.148 \\ 
Sparse4Dv2 & VoV-99 & \texttt{test}
& 0.556 & 0.638 & 0.462 & 0.238 & 0.328 & 0.264 & \textbf{0.115}\\ 
StreamPETR & ViT-L & \texttt{test}
& 0.620 & 0.676 &0.470  &0.241  & \textbf{0.258}  & 0.236  &0.134\\ 
\rowcolor[gray]{.9} 
\far{}~(Ours)  &  ViT-L & \texttt{test}
& \textbf{0.635} & \textbf{0.687} & \textbf{0.432} & \textbf{0.237} & 0.278 & \textbf{0.227} & 0.130 \\ 
\bottomrule
\end{tabular}}
\vspace{-0.3cm}
\end{table*}  

Next, 3D object queries interact with multi-scale sampled features from $\mathbf{F^{'}}$, according to the above 2D reference points $\mathbf{P}_q^{2d}$. In this way, diverse features from various vis and scales are aggregated into 3D queries by considering their relative importance.

\subsection{Range-modulated 3D Denoising} \label{sec:denoise}

3D object queries at different distances have different regression difficulties, which is different from 2D queries that are usually treated equally for existing 2D denoising methods such as DN-DETR~\cite{li2022dn}. 
The difficulty discrepancy comes from query density and error propagation. On the one hand, queries corresponding to distant objects are less matched compared to close ones. On the other hand, small errors of 2D proposals can be amplified when introducing 2D priors to 3D adaptive queries, illustrated in Fig.~\ref{fig:dn}, not to mention which effect increases along with object distance. 
As a result, some query proposals near GT boxes can be regarded as noisy candidates, whereas others with notable deviation should be negative ones.
Therefore we aim to recall those potential positive ones and directly reject solid negative ones, by developing a method called range-modulated 3D denoising.

Concretely, we construct noisy queries based on GT objects by simultaneously adding positive and negative groups. For both types, random noises are applied according to object positions and sizes to facilitate denoising learning in long-range perception. 
Formally, we define the position of noisy queries as:
\begin{equation}
    \mathbf{\Tilde{P}} = \mathbf{P}_{GT} + \alpha f_{p}(\mathbf{S}_{GT}) + (1-\alpha)f_{n}(\mathbf{P}_{GT})
\end{equation}
where $\alpha \in \{0, 1\}$ corresponds to the generation of negative and positive queries, respectively. $\mathbf{P}_{GT}, \mathbf{S}_{GT} \in\mathbb{R}^{3}$ represents 3D center $(x, y, z)$ and box scale $(w, l, h)$ of GT, and $\mathbf{\Tilde{P}}$ is noisy coordinates. We use functions $f_{p}$ and $f_{n}$ to encode position-aware noise for positive and negative samples.

For positive noisy samples, we set $f_{p}(\mathbf{S}_{GT})$ as a linear function of 3D box scale with a random variable. 
We incorporate the offset constraint within GT boxes to guide the model in accurately reconstructing the GT from positive queries, while ensuring clear distinction from surrounding adjacent boxes.
For negative samples, the offsets are supposed to be relevant to their position range, thus we propose several implementations. For some examples, $f_{n}(\mathbf{P}_{GT})$ can be in forms of $log(\mathbf{P}_{GT})$, $\lambda_2 \mathbf{P}_{GT}$ or $\sqrt{\mathbf{P}_{GT}}$. We show these attempts in Sec.~\ref{sec:ablate}. 
Moreover, multi-group samples are generated for each GT object to enhance query diversity. Each group comprises one positive sample and $K$ negative samples. This approach serves as an imitation of noisy positive candidates and false positive candidates during training.

\section{Experiment}

\begin{figure*}[t!]
\centering
\includegraphics[scale=0.174]{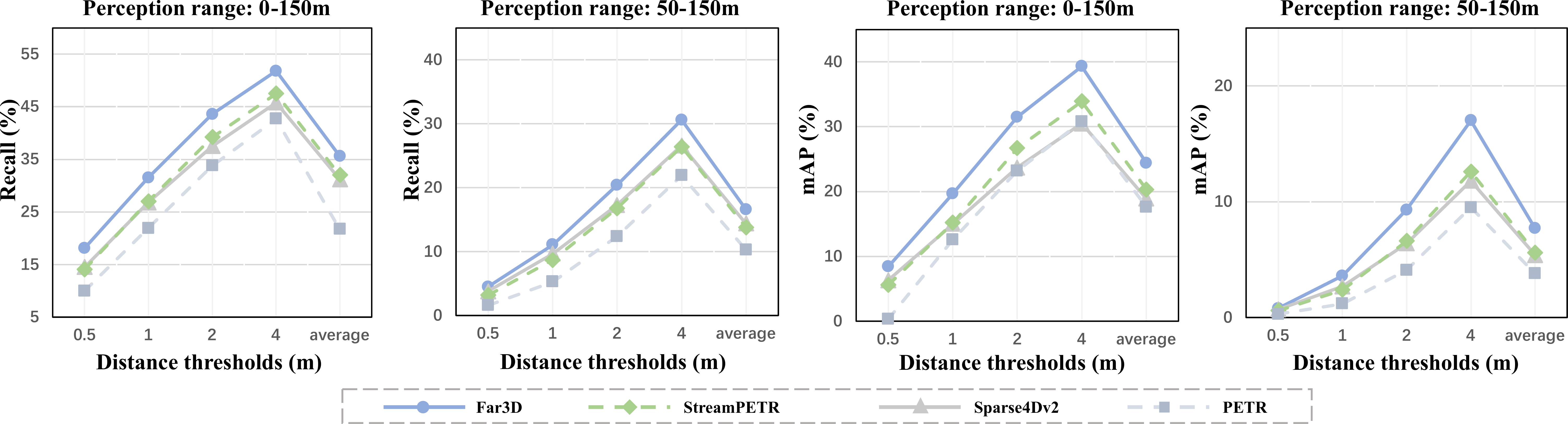}
\caption{3D Recall and AP of each method with different distance thresholds. Metrics of different ranges show that our approach consistently achieves a better result.}
\label{recall_ap}
\vspace{-0.3cm}
\end{figure*}

\subsection{Datasets and Metrics}
We use the large-scale Argoverse 2 dataset~\cite{wilson2023argoverse} and nuScenes dataset~\cite{caesar2020nuscenes} to explore and evaluate the effectiveness of our approach.

\textbf{Argoverse 2} is a dataset for perception and prediction studies in the autonomous driving domain. It contains 1000 scenes with 15 seconds duration and 10Hz annotation frequency for each scene. And these total scenes are divided into 700 for training, 150 for validation, and 150 for testing. Seven high-resolution ring cameras are provided with a combined 360° field of view. 
We evaluate it with 26 categories and a 150-meter range, satisfying the need for long-range tasks.
In addition to the mean Average Precision (mAP), we evaluate the methods with the metrics that Argoverse 2 dataset proposed: the Composite Detection Score (CDS), which is the main metric combining all factors in Argoverse 2 dataset, and three true positive metrics, including ATE, ASE, and AOE.

\textbf{nuScenes} is one of the most trustworthy datasets for multi-camera 3D object detection containing 1000 driving scenes in total. Each scene, approximately 20 seconds long, is annotated in 10 categories with 3D bounding boxes for sampled keyframes. We further conduct experiments on the dataset and compare the results with other methods using the following metrics, including mAP and the nuScenes Detection Score (NDS).

\begin{table}[t]
    
    \centering
    \caption{Ablation of our components on Argoverse 2 \texttt{val} set. StreamPETR is employed as the baseline, and we add the adaptive query, perspective-aware aggregation (PA) and range-modulated 3D denoising in order.}
    \vspace{-0.2cm}
    \label{tab:ablate_component}
    \tiny
    \resizebox{0.48\textwidth}{!}{
    \setlength{\tabcolsep}{2pt}
    \begin{tabular}{c|c|c|c|c|c}    
    \toprule
    \# & \textbf{Adaptive Query} & \textbf{PA} &
    \textbf{3D Denoising} & \textbf{mAP[\%]}$\uparrow$ & \textbf{CDS[\%]}$\uparrow$  \\  
    \toprule
    1  &     &     &       & 20.3    & 14.6           \\    
    2  & \ding{52}    &     &       & 22.4    & 16.1           \\
    3  & \ding{52}    & \ding{52}    &       & 23.4    & 17.3           \\
    \rowcolor[gray]{.9} 
    4  & \ding{52}    & \ding{52}    & \ding{52}      & \textbf{24.4} (+4.1)    & \textbf{18.1} (+3.5) \\
    \bottomrule
    \end{tabular}
    }
    \vspace{-0.3cm}
\end{table}
\subsection{Implementation Details}
With StreamPETR~\cite{wang2023exploring} as our baseline,
Far3D is composed of a backbone, an FPN neck, a 2D proposal head, and a 3D detection head. We adopt VoVNet-99~\cite{lee2019energy} pre-trained with FCOS3D~\cite{wang2021fcos3d} on nuScenes as the backbone to conduct main experiments. ViT-Large~\cite{dosovitskiy2020image} pre-trained by Objects365~\cite{shao2019objects365} and COCO~\cite{lin2014microsoft} dataset is used to scale up our model. By default, the FPN gives 4-level feature maps with sizes of 1/8, 1/16, 1/32, and 1/64. The perception range is set as 152.4m $\times$ 152.4m.

We use AdamW~\cite{loshchilov2017decoupled} optimizer with a weight decay of 0.01. The total batch size is 8 and the learning rate is set to 2e-4. 
The models are totally trained for 6 epochs, following the previous method~\cite{chen2023voxelnext}. Since the resolution of the front-view image is different from other views in Argoverse 2 dataset, we first resize the front image to a consistent resolution, then do the same image data augmentation as other images do. We do not use any BEV data augmentation on Argoverse 2 dataset. On the nuScenes dataset, we set the batch size as 32 and use the ResNet101~\cite{he2016resnet} backbone to train our method for 60 epochs. Other settings keep in line with StreamPETR.

\subsection{Main Results}
\noindent\textbf{Argoverse 2 Dataset.}
We compare the proposed framework with the existing state-of-the-arts on Argoverse 2 \texttt{val} set. As shown in Tab.~\ref{tab:av2_val_set}, when adopting VoV-99 backbone and 960$\times$640 input size, our method demonstrates a substantial superiority over other methods, achieving an impressive margin of 4.1\% mAP and 3.5\% CDS.
Besides the listed sparse query-based methods, we also conduct experiments on dense BEV-based methods, BEVStereo~\cite{li2022bevstereo} and SOLOFusion~\cite{park2022time}. The results are barely satisfactory and we suppose that is because of the greater difficulty of depth estimation. 
We also reproduce MV2D~\cite{wang2023object} but it can hardly converge here. The reason is mainly the generated anchors lack accurate depth estimation, leading to large localization deviations over long distances.
To sum up, the convergence problem in long-range detection is severe for the above methods, and we believe that our depth estimation and 3D denoising play key roles to solve it. More explanations are in the supplementary.

\begin{table}[t]
    \centering
    \caption{Ablation study with different score threshold $\tau$ for 2D proposals.}
    \vspace{-0.2cm}
    \label{tab:ablate_score}
    \tiny
    \resizebox{0.48\textwidth}{!}{
    \setlength{\tabcolsep}{3pt}
    \begin{tabular}{c|cc|ccc}  
    \toprule
    \textbf{$\tau$} & \textbf{mAP[\%]}$\uparrow$ & \textbf{CDS[\%]}$\uparrow$ & \textbf{mATE}$\downarrow$ & \textbf{mASE}$\downarrow$   &\textbf{mAOE}$\downarrow$ \\
    
    \midrule
    0.01 & 23.1 & 17.2 & 0.807 & 0.307 & 0.531 \\
    0.05 & 23.4 & 17.3 & 0.806 & 0.312 & 0.531 \\
    \rowcolor[gray]{.9} 
    0.1 & \textbf{24.4} & \textbf{18.1} & \textbf{0.796} & \textbf{0.304} & 0.538 \\
    0.2 & 23.7 & 17.6 & 0.802 & 0.307 & \textbf{0.530} \\
    0.3 & 23.5 & 17.4 & 0.799 & 0.307 & 0.577 \\
    \bottomrule
    \end{tabular}
    }
    \vspace{-0.3cm}
\end{table}
We further compare it with LiDAR-based SoTAs, CenterPoint~\cite{yin2021center}, FSD~\cite{fan2022fully}, and VoxelNeXt~\cite{chen2023voxelnext}. With a ViT-L backbone and 1536$\times$1536 resolution, our method outperforms them, showcasing the great potential of surround-view methods. In detail, LiDAR-based methods have a lower localization error (i.e. ATE) due to accurate depth information, while surround-view ones identify orientation properties (i.e. AOE) better. 

\begin{figure*}[t!]
\centering
\includegraphics[scale=0.167]{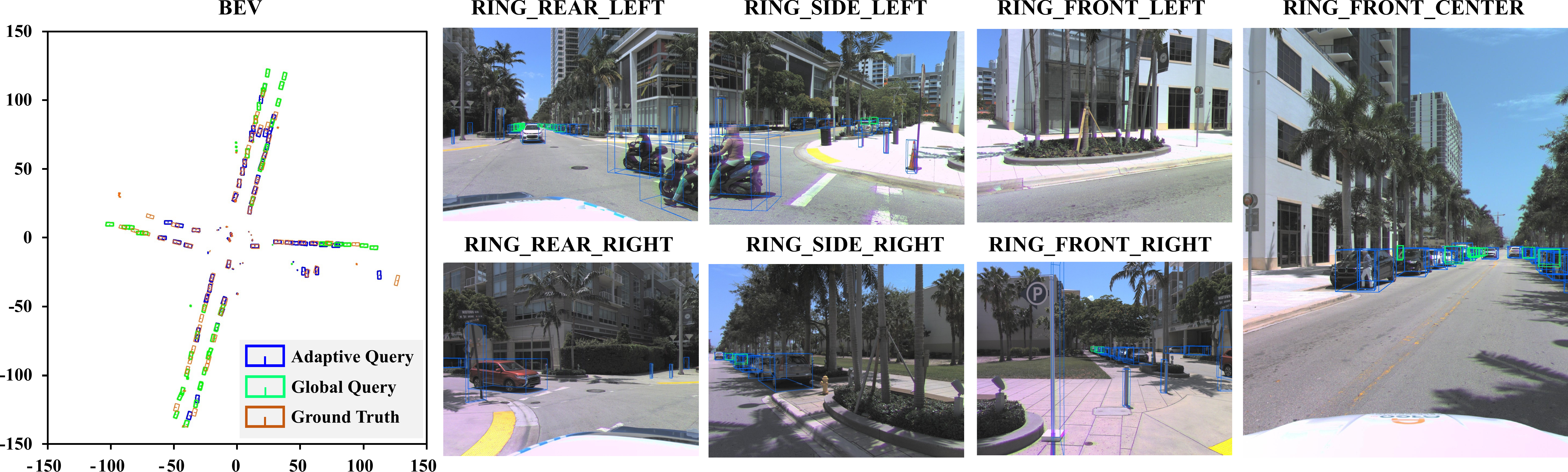}
\caption{Visualization results on Argoverse 2 dataset. 
We show 3D bounding boxes predicted both in multi-camera images and bird’s eye view. 
As illustrated, the view of the front center is distinguished from the other six views. The detection boxes predicted from 3D adaptive queries and 3D global queries are drawn in blue and green respectively. The GTs in orange are presented in BEV only.}
\vspace{-0.3cm}
\label{scene_vis}
\end{figure*}
As shown in Fig.~\ref{recall_ap}, we present the 3D recall and mAP results with different distances of 0-150m and 50-150m. Far3D consistently outperforms other methods. For distant objects, Far3D has a greater improvement when comparing recall and mAP with thresholds of 2m and 4m.

\noindent\textbf{nuScenes Dataset.}
To evaluate the generalization ability of our approach, we conducted additional comparisons on nuScenes dataset, as shown in Tab.~\ref{tab:nus_val_set}. Notably, our method outperforms previous SoTA methods with impressive results, achieving 51.0\% mAP and 59.4\% NDS on the \texttt{val} set and 63.5\% mAP and 68.7\% NDS on the \texttt{test} set. These superior metrics specifically highlight its effectiveness.


\subsection{Ablation Study \& Analysis} \label{sec:ablate}

In this section, we present a comprehensive analysis of the essential components of our model. As shown in Tab.~\ref{tab:ablate_component}, we start from StreamPETR as the baseline in \#1 and add each module to verify its effect. 


\noindent\textbf{Adaptive Query.}
Comparing \#1 and \#2 in Tab.~\ref{tab:ablate_component}, we can observe that adaptive query brings an improvement of 2.1\% mAP and 1.5\% CDS. Adaptive queries are insensitive to object range due to the robustness of 2D detectors in images, thus it is more suitable for general detection scenarios. 
To choose the optimal score threshold of 2D proposals, we conduct experiments shown in Tab.~\ref{tab:ablate_score}. Besides, we visualize the detection results in Fig.~\ref{scene_vis} and distinguish the boxes predicted from 3D adaptive queries and 3D global queries. The predictions from 3D adaptive queries cover a larger range, showing their indispensable significance.

\noindent\textbf{Perspective-aware Aggregation.} Adding the perspective-aware aggregation contributes a gain of 1.0\% mAP and 1.2\% CDS. Distant objects only occupy a few pixels on the image, therefore employing multi-level scales and views brings rich features according to different object locations. 

\begin{table}[t]
    \centering
    \caption{Performance Comparison of negative denoising samples with different designs and numbers.}
    \vspace{-0.2cm}
    \label{tab:ablate_dn}
    \tiny
    \resizebox{0.47\textwidth}{!}{
    \setlength{\tabcolsep}{5pt}
    \begin{tabular}{c|c|cc}  
    \toprule
    \textbf{\# Negative sample} & \textbf{Method} & \textbf{mAP[\%]}$\uparrow$ & \textbf{CDS[\%]}$\uparrow$   \\ 
    
    \midrule
    0 & -- & 23.4 & 17.3  \\ 
    1 & $log(\cdot)$ & 24.0 & 17.7 \\ 
    \rowcolor[gray]{.9} 
    2 & $log(\cdot)$ & \textbf{24.4} & \textbf{18.1}  \\ 
    3 & $log(\cdot)$ & 24.3 & 18.0 \\ 
    \midrule
    2 & $linear$ & 24.1 & 17.9 \\ 
    2 & $sqrt$ & 24.0 & 17.7 \\ 
    2 & $fixed$ & 23.7 & 17.6 \\ 
    \bottomrule
    \end{tabular}
    }
    \vspace{-0.3cm}
\end{table}
\noindent\textbf{Range-modulated 3D Denoising.} 3D denoising brings an improvement of 1.0\% mAP and 0.8\% CDS. Penalizing negative samples flexibly alleviates the challenge of false proposals and helps localize 3D objects, by taking the object range into consideration.  
We present experiments on different noising designs and numbers of negative samples, shown in Tab.~\ref{tab:ablate_dn}. The results imply that the logarithm function and two negative samples are optimal settings.

\begin{table}[t]
    \centering
    \caption{The impact of global query number. StreamPETR suffers from the convergence problem, where NaN denotes the failed training. In contrast, our framework shows robust performance even with only adaptive queries.}
    \vspace{-0.2cm}
    \label{tab:ablate_qnum}
    \tiny
    \resizebox{0.48\textwidth}{!}{
    \setlength{\tabcolsep}{3.5pt}
    \begin{tabular}{c|ccc|ccc}  
    \toprule
    \multirow{2}{*}{\textbf{\# Global query }} & \multicolumn{3}{c|}{\textbf{StreamPETR}} & \multicolumn{3}{c}{\textbf{\far{}~(Ours)}}  \\ 
    & 100 & 300 & 644 & 100 & 300 & 644 \\ 
    \toprule
    \textbf{mAP[\%]}$\uparrow$  & 1.5 & 16.9 & 20.5  & 23.5 & 23.6 & 24.4 \\ 
    \textbf{CDS[\%]}$\uparrow$ & 0.9 & 11.8 & 14.8  & 17.4 & 17.5 & 18.1 \\ 
    \bottomrule
    \end{tabular}
    }
    \vspace{-0.2cm}
\end{table}
\noindent\textbf{Effect of the Global Query.}
We also design the experiment to investigate the effect of global query in Tab.~\ref{tab:ablate_qnum}. 3D global queries and adaptive queries coexist in our framework and compensate for each other. As a baseline, StreamPETR suffers from the convergence problem when using a small number of global queries (e.g. 100), and only works for a sufficient amount. In contrast, our method showcases distinctive robustness. As the number of global queries decreases, our performance shows a slight decline. 

\section{Conclusion}
In this paper, we present a sparse query-based method for 3D long-range detection. Our approach incorporates 3D adaptive queries derived from 2D object priors, yielding high-quality proposals for the decoder. 
To improve training efficacy, we introduce a perspective-aware aggregation and range-modulated 3D denoising technique. Experimental results demonstrate the promising performance of our method, indicating its great potential for practical applications. 

\noindent\textbf{Limitations and Future Work.}
Despite our development for long-range detection, several limitations require future solutions. 
On the one hand, existing approaches exhibit poor performance on long-tail classes, ultimately lowering the average precision on Argoverse 2 dataset. On the other hand, evaluating long-range and close-range objects using unified metrics may not be suitable, emphasizing the need for practical and dynamic evaluation criteria that cater to diverse real-world scenarios.


\bibliography{aaai24}

\newpage
\section{Supplementary}
\noindent\textbf{The background of sparse query-based methods.} 
To localize 3D objects from surround-view images, DETR3D~\cite{wang2022detr3d} defines 3D object queries and generates 3D reference points, then samples 2D image features through coordinate projection and cross-attention. Finally, object queries are updated by aggregated features and used to predict the bounding boxes.
PETR~\cite{liu2022petr} generates 3D position-aware features by encoding the 3D coordinate information into position embedding. 
PETRv2~\cite{liu2022petrv2} develops multi-frame temporal modeling to boost 3D detection, and StreamPETR~\cite{wang2023exploring} takes a step further by proposing an object-centric temporal mechanism, which enables long-sequence query propagation and online prediction.

\noindent\textbf{Novelty of the model.} Our main contribution is the unified framework for long-range 3D detection, rather than stand-alone components. 
To tackle the poor 3D recall, heavy computation cost, and query error propagation existing in long-range detection, we introduce adaptive query, perspective-aware aggregation and 3D denoising strategies. 
Combining these strategies in an intuitive manner, we achieve the scalability of perception range. 


\noindent\textbf{Benefits in various ranges.} 
In fact, Far3D brings performance improvements in both close-range and long-range.
Empirically, we observe a common phenomenon of existing methods: the performance of close-range objects will decrease significantly when switching the training range from close range (e.g. 50m) to long range (e.g. 150m). 
Far3D mitigates the problem thanks to adaptive query and 3D denoising, thus it not only improves the performance of far objects, but also alleviates the performance degradation of near objects, as shown in Table~\ref{tab:ablate_distance}.

\begin{figure}
    \centering
    \includegraphics[scale=0.2]{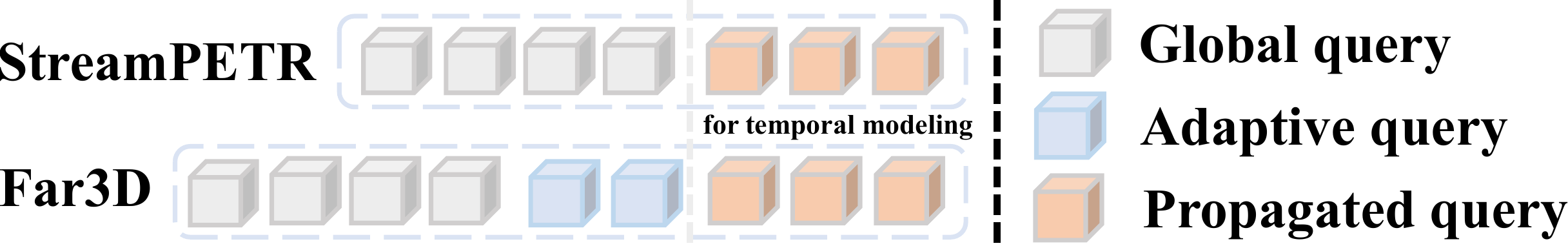}
    \caption{There are three types of queries in Far3D: global queries, adaptive queries and propagated queries. For temporal modeling, propagated queries are from previous frames with high-scores, the same as StreamPETR.
    }
    \label{fig:query}
\end{figure}

\noindent\textbf{Temporal modeling.} We employ propagated queries (depicted in Figure.~\ref{fig:query}) from previous frames to incorporate temporal features, following StreamPETR. Propagated queries are selected according to query score and irrelevant to original query type.

\noindent\textbf{Difference with recent methods.}
We highlight the distinctions between our approach and recent methods such as MV2D~\cite{wang2023object} and BEVFormer v2~\cite{yang2023bevformer}.
1) Motivations: First and foremost, our motivations differ significantly. Far3D focuses on tackling long-range detection challenges by leveraging 3D adaptive queries capable of adapting to dynamic scenarios and distant objects. In contrast, MV2D primarily aims to elevate 2D detectors to perform 3D detection in conventional detection tasks. BEVFormer v2 explores the synergy between image backbones and BEV detectors by incorporating perspective supervision.
2) Model Designs: Introducing 3D adaptive queries derived from 2D predictions equip the model with flexibility, yet it is not enough to tackle long-range detection. Experimentally, we found that another severe issue is the convergence problem. It is hard to converge for most existing methods, due to the impact of distant objects. Our proposed 3D denoising technique alleviates it significantly, facilitating the model convergence and performance.
3) Performance Superiority: Significantly, our proposed approach achieves exceptional performance compared to previous methods on the Argoverse 2 dataset. Notably, MV2D can hardly converge due to the considerable localization uncertainty associated with distant objects. Similarly, when evaluated on the popular nuScenes dataset, Far3D consistently surpasses MV2D (Far3D 51.0 mAP vs. MV2D 47.1 mAP on \texttt{val}) and BEVFormer v2 (Far3D 63.5 mAP vs. BEVFormer v2 55.6 mAP on \texttt{test}) by a significant margin.
In conclusion, the distinctiveness of Far3D lies in its \textbf{motivations}, \textbf{model designs}, as well as its \textbf{superior performance}.

\begin{table}[t]
    \centering
    \caption{Performance comparison in different perception ranges. 
    We train the StreamPETR and Far3D with 50m and 150m ranges respectively, and present the results of both in 0-50m and 50-150m. 
    Far3D alleviates the performance degradation in close range while improving the performance by larger scale in long range, compared to StreamPETR.
    }
    \vspace{-0.25cm}
    \label{tab:ablate_distance}
    \tiny
    \resizebox{0.48\textwidth}{!}{
    \setlength{\tabcolsep}{3.5pt}
    \begin{tabular}{c|c|cc|cc}  
    \toprule
    \multirow{2}{*}{\textbf{\# Perception range }} & \multirow{2}{*}{\textbf{train range }} & \multicolumn{2}{c|}{\textbf{test mAP[\%]}$\uparrow$} & \multicolumn{2}{c}{\textbf{test Recall[\%]}$\uparrow$}  \\ 
    & & 0-50m & 50-150m & 0-50m & 50-150m \\ 
    \toprule
    \textbf{StreamPETR} & 50m & 34.3 & 1.6 & 51.9 & 4.2 \\
    \textbf{StreamPETR} & 150m & 31.8(-2.5) & 7.4(+5.8) & 46.9(-5.0)  & 17.2(+13.0) \\
    \midrule 
    \textbf{Ours} & 50m & 38.7 & 1.9 & 55.0 & 4.6 \\
    \textbf{Ours} & 150m & \textbf{37.7}(-1.0) & \textbf{9.9}(+8.0) & \textbf{51.9}(-3.1) & \textbf{20.5}(+15.9) \\
    \bottomrule
    \end{tabular}
    }
    \vspace{-0.7cm}
\end{table}

\begin{figure*}[htb!]
  \centering
  \begin{subfigure}{0.4\textwidth}
    \centering
    \includegraphics[width=\textwidth]{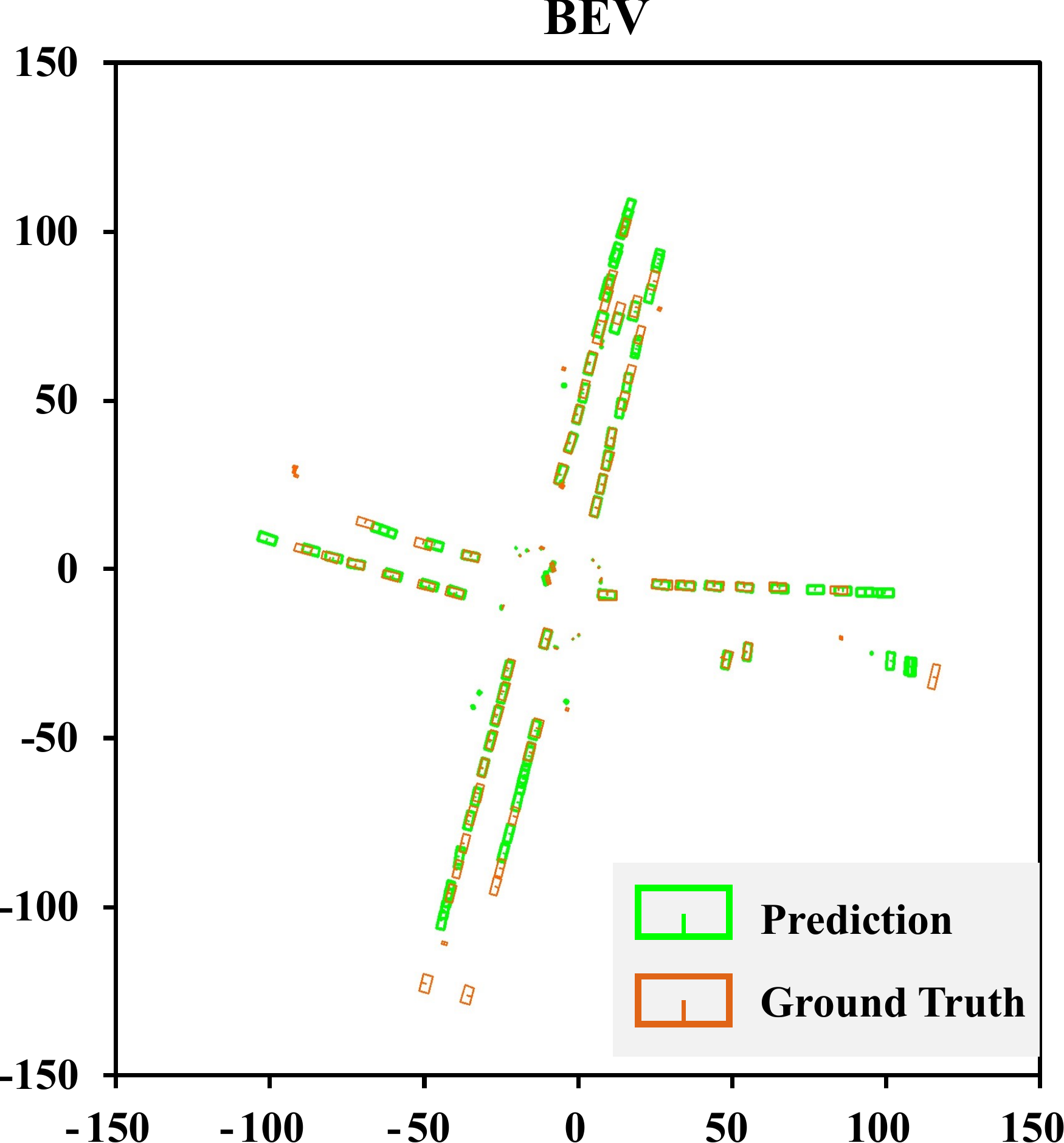}
    \caption{Far3D}
  \end{subfigure}
  \begin{subfigure}{0.4\textwidth}
    \centering
    \includegraphics[width=\textwidth]{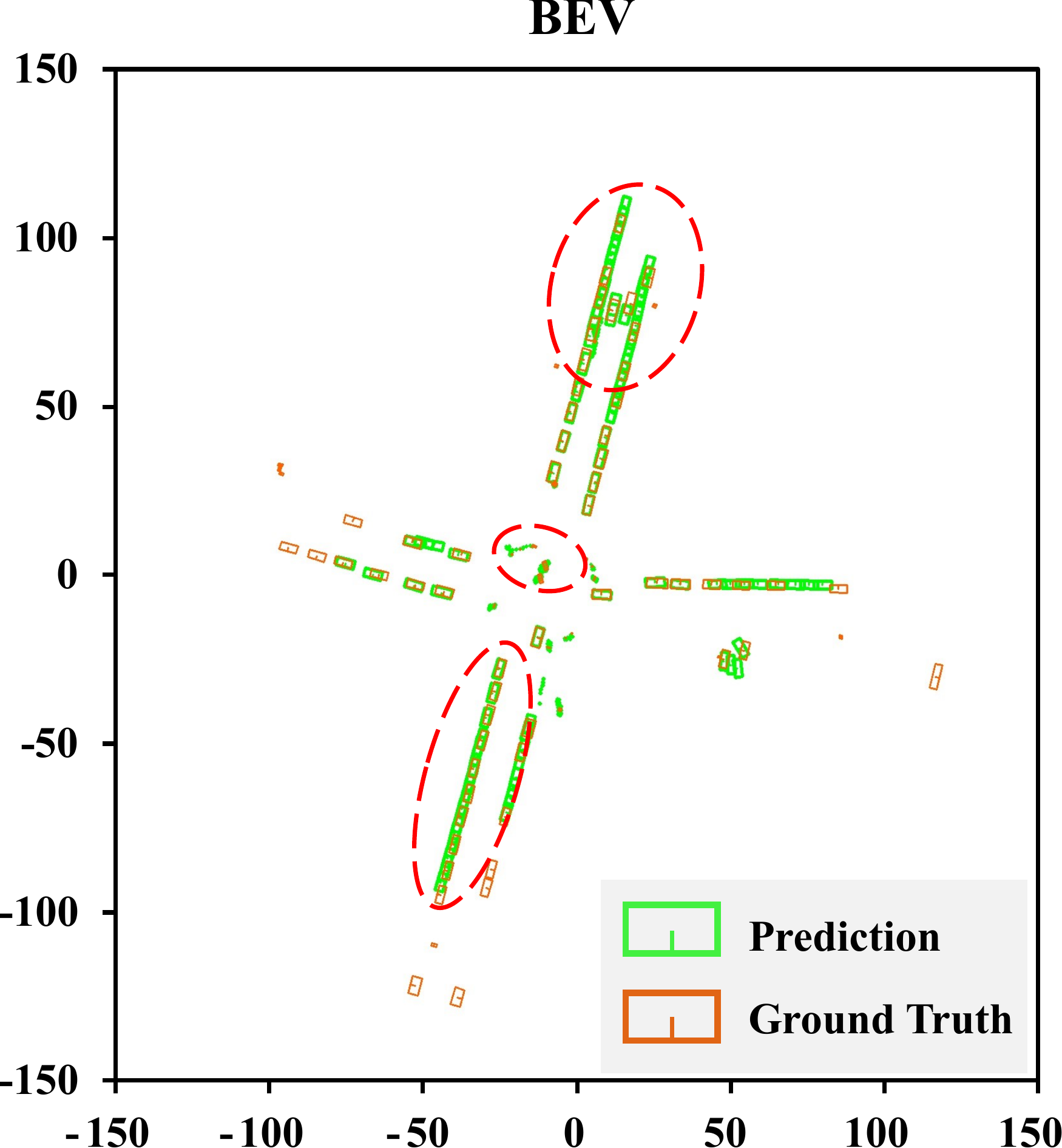}
    \caption{SOLOFusion}
  \end{subfigure}
  \caption{Visulization results of Far3D and SOLOFusion.}
  \label{fig:vis_compare}
\end{figure*}

Besides the above analysis, we also present visualization comparisons between Far3D and SOLOFusion~\cite{park2022time} in Fig.~\ref{fig:vis_compare}. SOLOFusion is a representative work of BEV-based methods. The visualizations revealed that SOLOFusion, even with NMS, generates numerous duplicate predictions, which led us to speculate that this issue may arise from the limited receptive field of the detection head when dealing with a large perception range.

\begin{figure*}[hbt!]
\centering
\includegraphics[width=\textwidth]{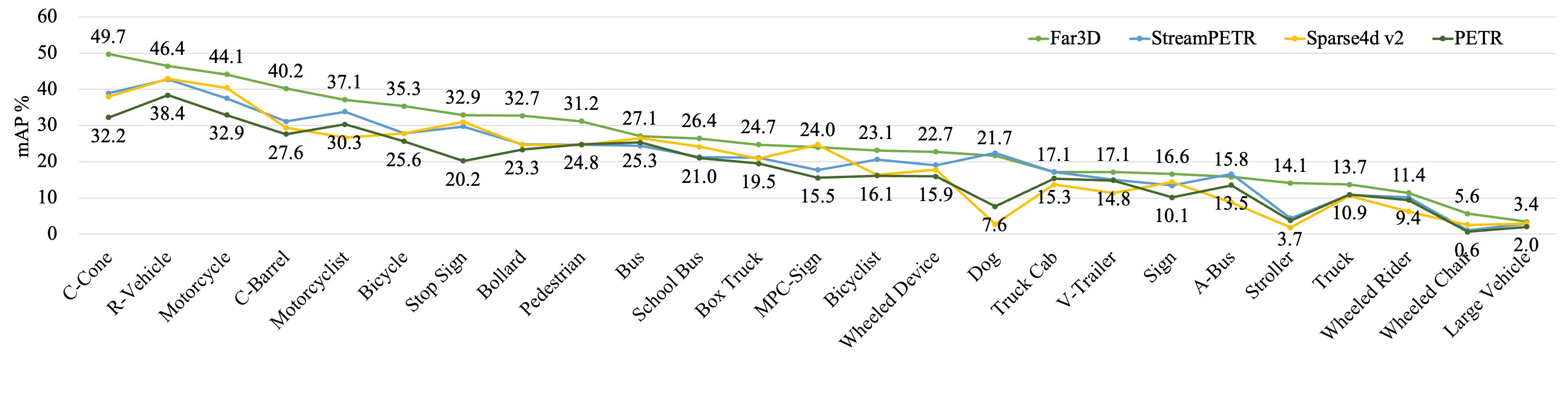}
\caption{Performance of all categories on Argoverse 2 \texttt{val} set, with adopting VoV-99 backbone. We discard Message Board Trailer due to its near-zero result. }
\label{exp_av2_cls}
\end{figure*}

\begin{figure*}[htb!]
  \centering
  \begin{subfigure}{0.49\textwidth}
    \centering
    \includegraphics[width=\textwidth]{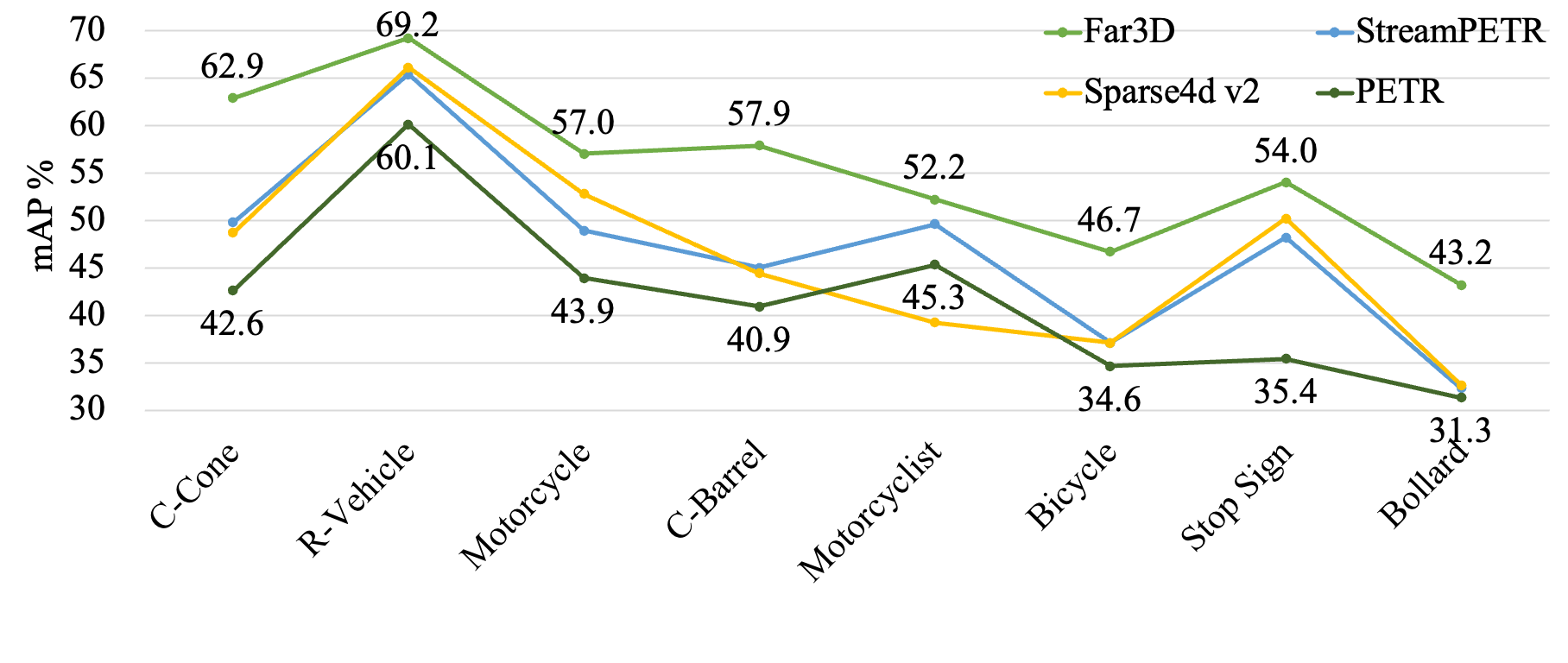}
    \caption{Range within 50 meters.}
  \end{subfigure}
  \hfill
  \begin{subfigure}{0.49\textwidth}
    \centering
    \includegraphics[width=\textwidth]{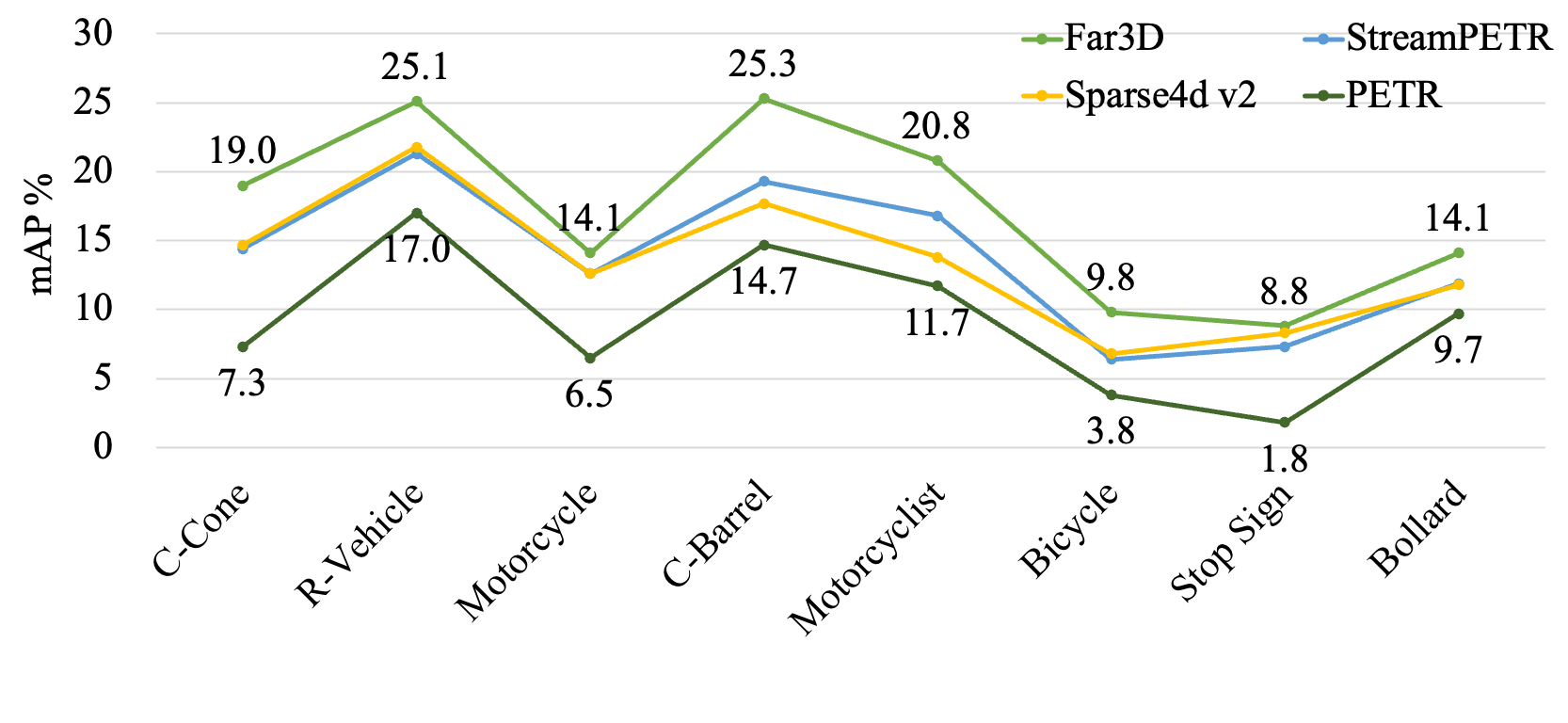}
    \caption{Range 50-100 meters.}
  \end{subfigure}
  \caption{Performance of first eight categories in different ranges.}
  \label{fig:cls_range}
\end{figure*}

\noindent\textbf{Performance comparisons of all categories.} 
There are 26 categories in Argoverse 2, far more than other datasets. Fig.~\ref{exp_av2_cls} shows the performance comparisons in detail. Furthermore, results of the range 0-50m and 50-100m are presented in Fig.~\ref{fig:cls_range}. Our method consistently achieves the best results.

\noindent\textbf{The statistics of adaptive queries.}
For a deeper analysis of the superiority of adaptive queries, we make statistics during training. We observe that the number of adaptive queries of Argoverse 2 dataset is 92 on average for each sample, with a maximum of 236 and a minimum of 11, accounting for only a small proportion of the total. We make an experiment that uses extra 92 queries instead of the adaptive ones, leading to a decrease of 1.5\% mAP and 1.3\% mCDS. The result validates the distinctive contribution of adaptive queries.

\noindent\textbf{More Details of Far3D.}
We provide additional details about our proposed Far3D as follows:
1) Our adaptive queries are generated by transforming 2D proposals and corresponding depth estimates into 3D space. To ensure the training efficacy, during the early stages of training, we utilize ground truth (GT) depth to generate 3D adaptive queries. As the network training stabilizes, we introduce predicted depth for the adaptation process. 
2) Considering long-range tasks, the image pixel areas occupied by 3D objects at different ranges exhibit significant variation. A single-scale feature representation alone may not address the diverse requirements of different queries in 3D detection. To tackle this, we incorporate multi-scale features (p2-p5) obtained from the FPN. The query undergoes iterative updates using a deformable attention mechanism, reducing computational complexity. Our experiments indicate that the network can adaptively match objects at different distances and leverage multi-scale features effectively. We have compared the approach that manually selects feature layers based on object distance, and the results align closely with that learned by the network.

\end{document}